\documentclass{article}

\usepackage[preprint]{neurips_2025}

\usepackage[utf8]{inputenc}    
\usepackage[T1]{fontenc}       
\usepackage{times}             
\usepackage{inconsolata}       

\PassOptionsToPackage{margin=1in}{geometry}
\usepackage{microtype}         
\usepackage{nicefrac}          

\usepackage{amsmath}
\usepackage{amssymb}
\usepackage{amsfonts}          
\usepackage{latexsym}          
\usepackage{dsfont}            

\usepackage{booktabs}          
\usepackage{tabularx}
\usepackage{multirow}
\usepackage{arydshln}          
\usepackage{makecell}          

\usepackage{algorithm}
\usepackage{algorithmicx}
\usepackage{algpseudocode}

\usepackage{hyperref}          
\usepackage{url}               

\usepackage{graphicx}
\usepackage{xcolor}
\usepackage{subcaption}
\usepackage{wrapfig}

\usepackage{listings}          
\usepackage{tcolorbox}         

\usepackage{newfloat}          
\usepackage{markdown}          
\usepackage{lineno}            

\usepackage{enumitem}          
\title{EL4NER: Ensemble Learning \\for Named Entity Recognition \\via Multiple Small-Parameter Large Language Models}

\author{
  Yuzhen Xiao$^{1,2}$\thanks{Equal contribution.} \\
  \texttt{xiaoyuzhen@stu.pku.edu.cn} \\
  \And
  Jiahe Song$^{1,2}$\footnotemark[1] \\
  \texttt{songjh@stu.pku.edu.cn} \\
  \And
  Yongxin Xu$^{1,2}$\footnotemark[1] \\
  \texttt{xuyx@stu.pku.edu.cn} \\
  \And
  Ruizhe Zhang$^{1,2}$ \\
  \texttt{nostradamus@stu.pku.edu.cn} \\
  \And
  Yiqi Xiao$^{4}$ \\
  \texttt{xiaoyiqi@buaa.edu.cn} \\
  \And
  Xin Lu$^{1}$ \\
  \texttt{luxin@stu.pku.edu.cn} \\
  \And
  Runchuan Zhu$^{1,2}$ \\
  \texttt{2201210572@stu.pku.edu.cn} \\
  \And
  Bowen Jiang$^{1,2}$ \\
  \texttt{2301210261@stu.pku.edu.cn} \\
  \And
  Junfeng Zhao$^{1,2,3}$\thanks{Corresponding author.} \\
  \texttt{zhaojf@pku.edu.cn} \\
  \\
  \small
  $^{1}$ School of Computer Science and School of Software \& Microelectronics, Peking University, Beijing, China\\
  \small
   $^{2}$ Key Laboratory of High Confidence Software Technologies, Ministry of Education, Beijing, China\\ 
  \small
   $^{3}$ Nanhu Laboratory, Jiaxing, China\\
   \small
   $^{4}$ School of Computer Science and Engineering, Beijing University of Aeronautics and Astronautics, Beijing, China
}

\begin{document}

\maketitle

\begin{abstract}
In-Context Learning (ICL) technique based on Large Language Models (LLMs) has gained prominence in Named Entity Recognition (NER) tasks for its lower computing resource consumption, less manual labeling overhead, and stronger generalizability. Nevertheless, most ICL-based NER methods depend on large-parameter LLMs: the open-source models demand substantial computational resources for deployment and inference, while the closed-source ones incur high API costs, raise data-privacy concerns, and hinder community collaboration. To address this question, we propose an \textbf{\underline{E}}nsemble \textbf{\underline{L}}earning Method \textbf{\underline{for}} \textbf{\underline{N}}amed \textbf{\underline{E}}ntity \textbf{\underline{R}}ecognition (EL4NER), which aims at aggregating the ICL outputs of multiple open-source, small-parameter LLMs to enhance overall performance in NER tasks at less deployment and inference cost. Specifically, our method comprises three key components. First, we design a task decomposition-based pipeline that facilitates deep, multi-stage ensemble learning. Second, we introduce a novel span-level sentence similarity algorithm to establish an ICL demonstration retrieval mechanism better suited for NER tasks. Third, we incorporate a self-validation mechanism to mitigate the noise introduced during the ensemble process. We evaluated EL4NER on multiple widely adopted NER datasets from diverse domains. Our experimental results indicate that EL4NER surpasses most closed-source, large-parameter LLM-based methods at a lower parameter cost and even attains state-of-the-art (SOTA) performance among ICL-based methods on certain datasets. These results show the parameter efficiency of EL4NER and underscore the feasibility of employing open-source, small-parameter LLMs within the ICL paradigm for NER tasks.
\end{abstract}

\section{Introduction}
Named Entity Recognition (NER) is a fundamental and crucial task in Natural Language Processing (NLP) \citep{nermeaning}, aimed at extracting structured data from unstructured texts while supporting downstream tasks such as knowledge graph construction \citep{kgner}, knowledge reasoning \citep{krner}, and question answering \citep{qaner}, among others. Recently, Large Language Models (LLMs) have demonstrated remarkable performance on NER tasks, attributed to their exceptional capability for generalization and language understanding \citep{du2022glm,brown2020language,touvron2023llama}.

Existing NER methods utilizing LLMs can be primarily divided into two categories: Supervised Fine-Tuning (SFT-based) and In-Context Learning (ICL-based) methods. In contrast to SFT-based methods, ICL-based methods do not require additional training on domain-specific datasets and can be readily adapted to new domains by simply adjusting prompts and demonstrations, which offer lower computing resource consumption, less manual labeling overhead and stronger generalizability \citep{llmsurvey,nersurvey}, gradually becoming widely adopted mainstream solutions. However, there are also some disadvantages in existing ICL-based methods. Due to the high demands on models' basic capabilities, ICL-based methods typicall resort to utilizing large-parameter LLMs, which provide strong contextual understanding and broad domain knowledge. Nevertheless, the dependence on large-parameter LLMs imposes a number of limitations on the practical application of such methods. On the one hand, open-source, large-parameter LLMs (such as DeepSeek-R1 \citep{guo2025deepseek}) demand high computational resource overhead for both deployment and inference. On the other hand, close-source, large-parameter LLMs (such as models from the GPT series \citep{hurst2024gpt}) bring high API call costs, worrying data privacy risks and limited community collaboration issues. Specifically, most previous ICL-based NER methods adopt the large-parameter LLMs from GPT series.

Motivated by the high costs and practical barriers of relying solely on large-parameter LLMs, we turn to small-parameter LLMs as a potentially more efficient alternative. Single small-parameter LLM without fine-tuning often struggles to recognize entities that require more domain knowledge to judge due to its limited knowledge stock. However, different small-parameter LLMs have unique strengths in various domains and can complement each other. This observation raises a critical question: Whether it is possible to leverage multiple open-source, small-parameter LLMs to achieve better performance at less deployment and inference cost? Inspired by ensemble learning theory \citep{llmblender}, we consider integrating multiple open-source small-parameter LLMs to meet or even exceed the performance of large-parameter LLMs in the NER task. 

\begin{figure*}[t]
    \centering
    \includegraphics[width=\textwidth]{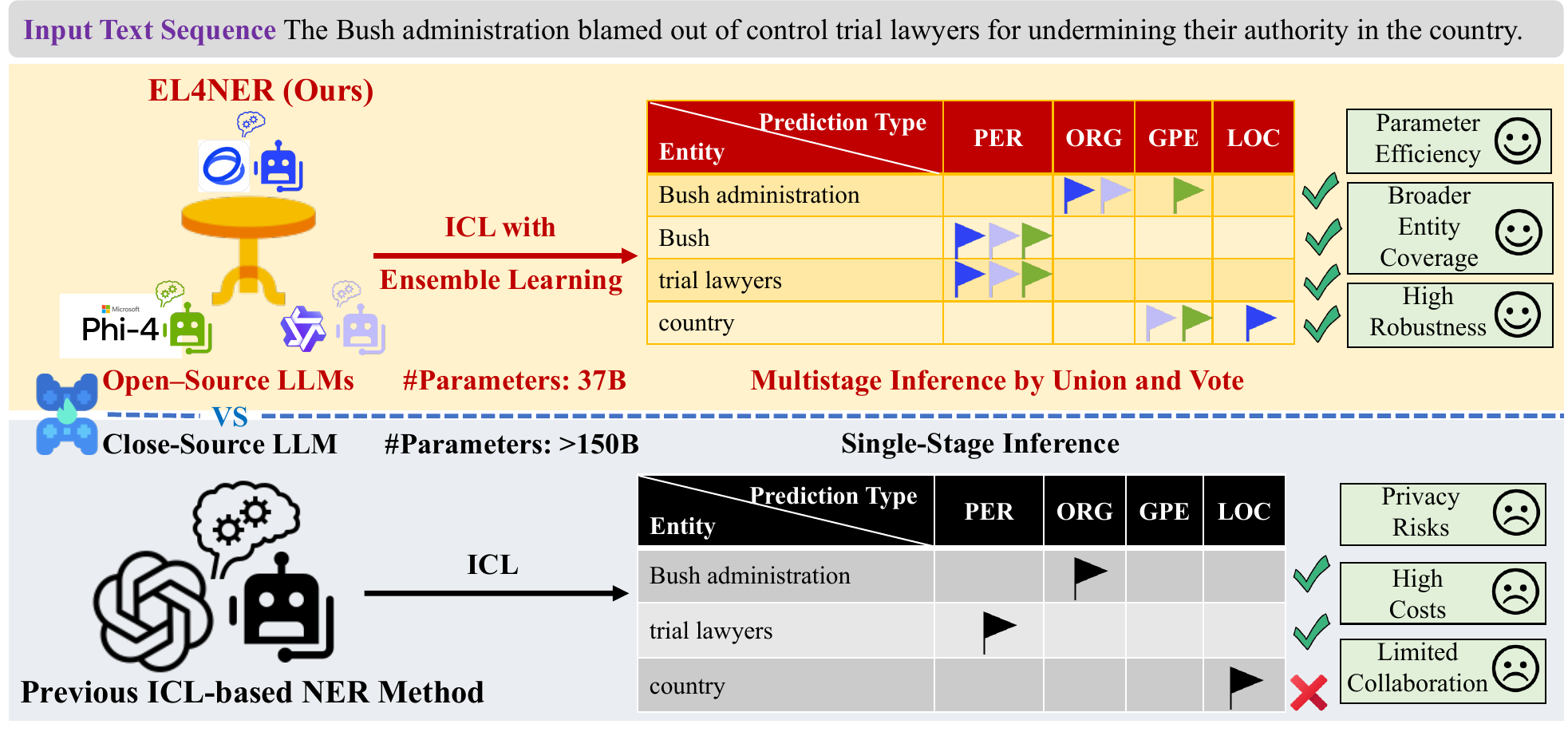}
    \caption{Illustration for the comparison of EL4NER with previous ICL-based methods. Most of the previous methods use closed-source large-parameter LLMs to perform single-stage ICL to accomplish the NER task, whereas EL4NER employs multiple open-source small-parameter LLMs to perform multi-stage ICL through ensemble learning to integrate the inference results of these LLMs by means of union and vote.}
    \label{fig:teaser}
\end{figure*}

Although seemingly straightforward, implementing this idea faces these challenges: \textbf{(C1)} Achieving desired results by merely integrating the final NER outputs of LLMs can be challenging. Therefore, how to enable a deep, multi-stage integration process? \textbf{(C2)} How to better energize the NER capacity of small-parameter LLMs in ICL process? \textbf{(C3)} Ensemble learning may introduce some noise into the NER results, so how can they be effectively filtered out? In response to these challenges, we propose EL4NER, an \textbf{\underline{E}}nsemble \textbf{\underline{L}}earning Method \textbf{\underline{for}} \textbf{\underline{N}}amed \textbf{\underline{E}}ntity \textbf{\underline{R}}ecognition based on ICL as shown in Fig. \ref{fig:teaser}. Specifically, for \textbf{C1}, EL4NER decomposes the NER task into two stages, the first of which uses multiple LLMs to extract potential entity spans and integrate them, and the second of which confirms entity types one by one by voting on multiple LLMs, thus enabling two integrations at a fine-grained level. For \textbf{C2}, we design a demonstration retrieval mechanism based on pre-extracting for important spans and weighting for part-of-speech tags to enhance the ICL effect. For \textbf{C3}, we introduce a self-validation mechanism to filter the noise brought by the ensemble.

Our main contributions are summarized as follows:
\begin{itemize}[leftmargin=*]
	\item Considering the characteristics of the NER task, we design a deep-level and multi-stage ensemble learning pipeline based on task decomposition.
	\item We propose an ICL demonstration retrieval mechanism tailored for the NER task, which calculates similarity between sentences from span-level based on span pre-extraction and weights of part-of-speech tags, significantly improving the effectiveness of ICL.
    \item Wide range of experiments on multi-domain datasets demonstrate the effectiveness of the proposed method, where EL4NER realizes excellent NER effect at a lower parameter cost, outperforms most of NER methods based on large-parameter LLMs, and achieves the state-of-the-art (SOTA) performance on several commonly used NER datasets among ICL-based methods.
\end{itemize}

\section{Related work}
\subsection{Zero-shot named entity recognition via LLMs}
LLMs have shown impressive performance in the field of named entity recognition (NER), especially their excellent generalization performance makes them substantially outperform traditional knowledge extraction models in zero-shot scenarios \citep{llmsurvey,chatie,gptner,SelfImprovingFZ}. Specifically, zero-shot learning paradigm is subdivided into Cross-domain learning and Zero-shot prompting. Cross-domain learning leverages datasets from domains that are not represented in the test sets for training purposes \citep{Wang2023InstructUIEMI,Sainz2023GoLLIEAG,Zhou2023UniversalNERTD}. In contrast, Zero-shot prompting entirely eliminates the training phase, relying solely on a variety of prompting techniques and combinations of multi-process methods for NER task, which is the focus of our study. GPT-NER \citep{gptner} proposes a basic paradigm for NER prompt and adds a self-verification mechanism, which is the first work to attepmt LLMs on NER task. Based on which, \citet{chatie,Xie2023EmpiricalSO,SelfImprovingFZ,picl,kim2024verifinerverificationaugmentednerknowledgegrounded,yan2024ltnerlargelanguagemodel} etc,. proposed different tricks to improve the performance of LLMs, such as using code format to restrict the I/O of LLMs, multi-stages decomposition, or various retrieval methods and so on (which will be detailed described in our baseline descriptions in Appendix~\ref{sec:Baseline Details}). However, the methods mentioned above overlook biases arising from training corpus, source, and methods when relying on single, large-parameter models such as GPT. In contrast, our framework leverages diverse model characteristics to enhance comprehensive checking, significantly reducing computational costs by utilizing lighter models with a total parameter count below 40B.
\subsection{Ensemble learning in LLMs}
Ensemble learning refers to the generation and combination of multiple inducers to solve a particular machine learning task. The main idea is that weighing and aggregating several individual opinions will be better than choosing the opinion of one individual \citep{ensemblelearningsurvey1,ensemble2,ensemblelearning}. With the popularity of LLMs, it became popular to use LLMs for ensemble learning. LLM-Blender \citep{llmblender} first proposes a rank-and-integrate pipeline framework for ensembling LLMs and test on MixInstruct which is a new dataset to benchmark ensemble models for LLMs in instruction-following tasks. Specifically, our work falls under the category of non-cascade unsupervised ensemble after inference \citep{chen2025harnessingmultiplelargelanguage}. Further, such approaches can be classified into two types: selection-based, which focuses on selecting a single response from multiple candidates \citep{li2024agentsneed,guha2024smoothielabelfreelanguage,sigetting}, and selection-then-regeneration \citep{tekin-etal-2024-llm,lv-etal-2024-urg,llmblender}, where a subset of candidate responses is initially selected and then fed into a generative model for regeneration to produce the final output. Different from the above methods, our framework for NER tasks accepts the full results of multiple LLMs to improve recall as much as possible, and then uses mechanisms such as self-verification to improve precision.

\section{Methdology}
\subsection{Task formulation}
\label{sec:form}
Given a text sequence $X$, the NER task is to extract $Y=\{y_{i}\}_{i=1}^{|Y|}$ from $X$, where $y$ represents a certain named entity. A named entity can be further represented as $y=(s,t),t\in\mathcal{T}$, where $s$ represents the text span of a entity, $t$ represents the type of this entity, and $\mathcal{T}$ represents the predefined set of entity types. 

We employ ensemble learning to extract $Y$ from $X$. For this purpose, we introduce the set of LLMs $\mathcal{M}=\{M_i\}_{i=1}^{|\mathcal{M}|}$, where $M$ represents a certain LLM. The LLMs in $\mathcal{M}$ will be used as backbones for inference by ICL, and their inference results will be integrated to obtain the final result. In EL4NER, the ways of ensemble include taking the union of the inference results from multiple LLMs and voting for the results.

ICL is usually categorized into zero-shot learning and few-shot learning. Zero-shot learning only inputs instructions to LLM, while few-shot learning tends to select some samples as demonstrations from a labeled candidate set $\mathcal{C}=\{(X^c_i,Y^c_i)\}_{i=1}^{|\mathcal{C}|}$, where a certain demonstration consists of $X^c$ and $Y^c$, which respectively represent a certain candidate text sequence and its entity label.
\subsection{Overview}

\begin{figure*}[t]
    \centering
    \includegraphics[width=\textwidth]{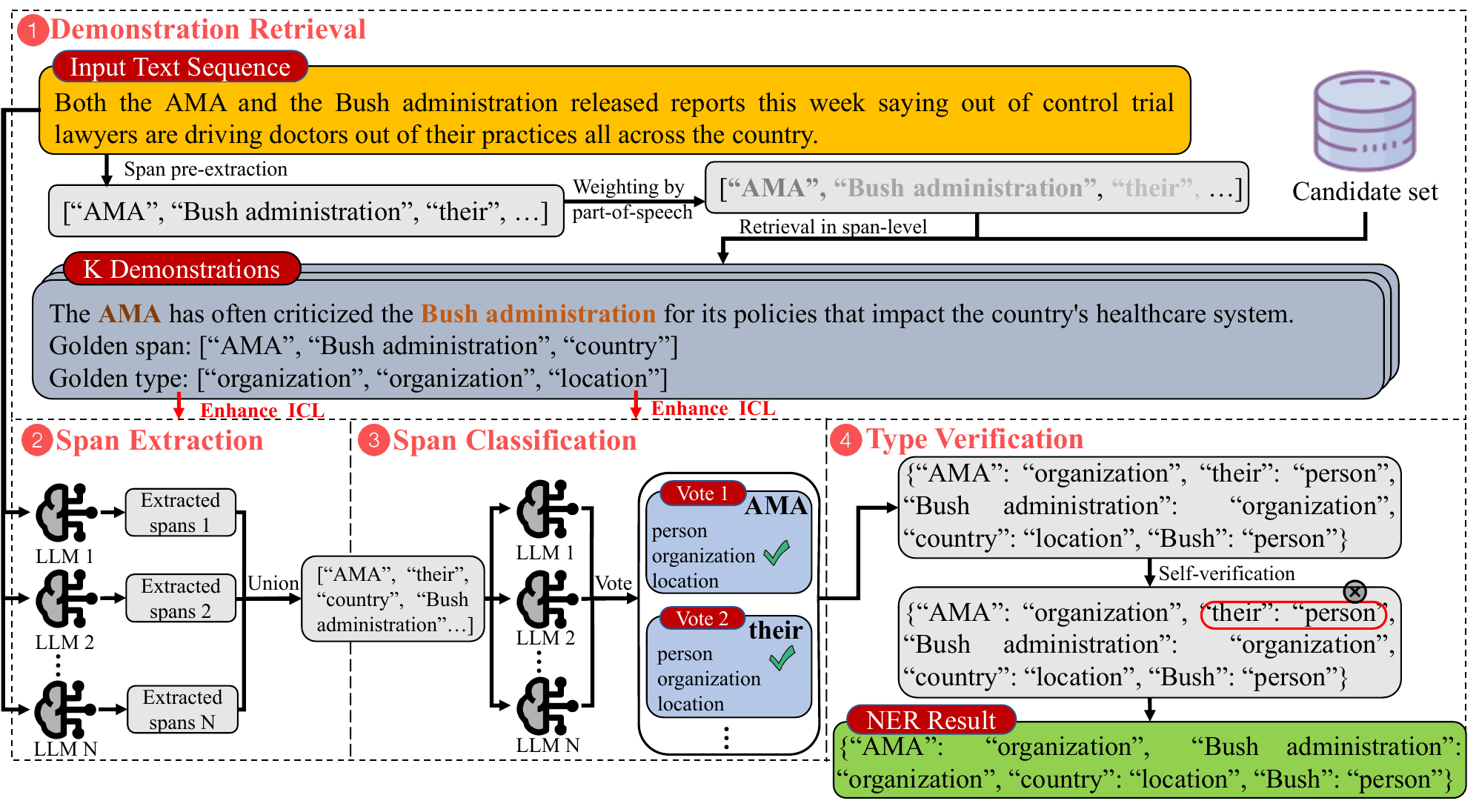}
    \caption{
Overview of the proposed EL4NER, which adopts a multi-stage ensemble learning method for NER. It includes four stages: (1) \textbf{Demonstration Retrieval}, selecting span-relevant samples for prompt construction; (2) \textbf{Span Extraction}, identifying potential entity spans; (3) \textbf{Span Classification}, assigning entity types to the extracted spans; (4) \textbf{Type Verification}, filtering out incorrectly classified spans. Each stage leverages multiple small-parameter LLMs to enhance robustness and effect of ICL.
}
    \label{fig:pipeline}
\end{figure*}

To integrate the inference results of backbones at a fine-grained level, we adopt the idea of task decomposition, which decomposes the NER task into two atomic tasks, span extraction and span classification, corresponding to the two stages in the pipeline of our method. ICL paradigm is employed in both span extraction and span classification stages, for which we include the demonstrations retrieved by our designed algorithm to improve its effect further. Fig. \ref{fig:pipeline} outlines the four stages in the pipeline of our method: \textbf{Demonstration retrieval} (Section \ref{sec:retrieval}), \textbf{Span extraction} (Section \ref{sec:extraction}), \textbf{Span classification} (Section \ref{sec:classification}), and \textbf{Type verification} (Section \ref{sec:verification}).

\subsection{Demonstration retrieval}
\label{sec:retrieval}
Since the outputs of NER are essentially some spans with special meaning, the demonstration retrieval stage aims at retrieving several demonstrations, of which the candidate text sequences are similar to the input text sequence at the span-level, to enhance the effect of ICL in the subsequent stages. As shown in Fig. \ref{fig:pipeline}, this stage can be overall divided into three steps: span pre-extraction, weighting by part-of-speech tags and retrieval in the span-level.

Firstly, in order to obtain the features of the input text sequence at the span-level, we roughly pre-extract the potential entity spans from the input text sequence by zero-shot learning. As shown in Eq. \ref{eq:pre-extraction}, we input $X$ and $P(I_{\text{ext}})$ to $M_i$ to get the set of spans pre-extracted by a single LLM via zero-shot learning, where $P(I_{\text{ext}})$ is the corresponding prompt consisting mainly of $I_{\text{ext}}$ (the instruction using to extract spans). Notably, we employ ensemble learning in this step. As shown in Eq. \ref{eq:pre-extraction_union}, we take the union of the span sets pre-extracted by the LLMs in $\mathcal{M}$ to obtain the final pre-extracted span set. For a candidate text sequence in the candidate set, we can adopt the same process as above to get its pre-extracted span set $\hat{S}^c$.

\begin{equation}
\label{eq:pre-extraction}
\hat{S_{i}}=M_{i}(X,P(I_{\text{ext}}))
\end{equation}
\begin{equation}
\label{eq:pre-extraction_union}
\hat{S}=\bigcup\limits_{i=1}^{|\mathcal{M}|}\hat{S_i}
\end{equation}

Secondly, the spans in $\hat{S}$ usually have different importance. The higher the likelihood that a span is an entity, the higher its importance. Typically, a proper noun has the highest likelihood of being a named entity, followed by a common noun and then a pronoun. In order to distinguish the importance between the different spans, we weight them according to the part-of-speech of their head word. The weighting function is as follows:
\begin{equation}
    w(s) = 
    \begin{cases}
        2^2, & \text{if } \mathrm{POS}(s)=\text{"PRON"} \\
        2^1, & \text{if } \mathrm{POS}(s)=\text{"NOUN"} \\
        2^0, & \text{if } \mathrm{POS}(s)=\text{"PROPN"} \\
        0, & \text{if } \mathrm{POS}(s)=\text{"OTHERS"} 
    \end{cases},
\end{equation}
where $s$ represents a certain span in $\hat{S}$ and $\mathrm{POS}$ represents a mapping from a span to part-of-speech of its head word, $\text{"PROPN"}$ represents proper noun, $\text{"NOUN"}$ represents common noun, $\text{"PRON"}$ represents pronoun, and $\text{"OTHERS"}$ represents other types for part-of-speech (which we don't think matters).

Thirdly, in order to retrieve demonstrations that are similar to the input text sequence at the span-level, we propose a span similarity  between the input text sequence and a candidate text sequence in the candidate set. For any span $s_x$ in $\hat{S}$, we compute the semantic similarity between it and each span $s_c$ in $\hat{S}^c$, and the span in $\hat{S}^c$ that is most similar to $s_x$ is called the matched span of $s_x$. For all spans $s_x$ in $\hat{S}$, we use $w(s_x)$ to weight and sum the semantic similarity between them and their matched spans (i.e. $\max\left(\mathrm{Sim}(s_x, s_c)\right)$) to obtain span similarity as follows:
\begin{equation}
\mathrm{SpanSim}(\hat{S}, \hat{S}^c)=\frac{\sum\limits_{s_{x}\in \hat{S}} w(s_x) \max\limits_{s_c \in \hat{S}^c} \left(\mathrm{Sim}(s_x, s_c)\right)}{\sum\limits_{s_x\in\hat{S}}w(s_x)}, 
\end{equation}
where $\mathrm{SpanSim}$ represents the span similarity function, and $\mathrm{Sim}$ represents the semantic similarity function. By calculating the span similarity between $X$ and each candidate text sequence in the candidate set, we can retrieve the top-k similar samples to form the demonstration set $D$.

\subsection{Span extraction}
\label{sec:extraction}
As shown in Fig. \ref{fig:pipeline}, after retrieving the demonstrations, the span extraction stage is aimed at extracting potential entity spans from the input text sequence again by few-shot learning, which usually leads to broader entity
coverage than zero-shot learning in span pre-extraction. As shown in Eq. \ref{eq:extraction} and Eq. \ref{eq:extraction_union}, the process of span extraction is similar to span pre-extraction. The difference between the two is that the prompt $P(I_{\text{ext}},D_{\text{ext}})$ used for span extraction contains demonstrations $D_{\text{ext}}$, which is obtained by collating the candidate text sequences and their golden span sets from $D$. Finally, we can obtain the integrated span set $S$ by ensemble learning.
\begin{equation}
\label{eq:extraction}
S_{i}=M_{i}(X,P(I_{\text{ext}},D_{\text{ext}}))
\end{equation}
\begin{equation}
\label{eq:extraction_union}
S=\bigcup\limits_{i=1}^{|\mathcal{M}|}S_i
\end{equation}

\subsection{Span classification}
\label{sec:classification}
As shown in Fig. \ref{fig:pipeline}, the span classification stage is aimed at performing robust type judgment by vote from multiple LLMs.

We hope to reduce the variance of a single LLM's decision and obtain more robust classification results by having multiple LLMs to jointly decide the type of each potential entity span in a voting manner. As shown in Eq. \ref{eq:classfication}, we input $X$, $S$, and $P(I_{\text{cls}},D_{\text{cls}})$ to $M_i$ to obtain a single LLM's classification result set $T_i$ for spans in $S$ by few-shot learning, where $P(I_{\text{cls}},D_{\text{cls}})$ is the corresponding prompt consisting mainly of $I_{\text{cls}}$ (the instruction using to predict the types of the spans) and $D_{\text{cls}}$ (the demonstrations set obtained by collating candidate text sequences, their golden span sets, and their golden type sets from $D$), and every prediction type in $T_i$ belongs to $\mathcal{T}$. As shown in Eq. \ref{eq:vote}, we use hard voting to integrate the classification results $T_i (i=1,2,\ldots,|\mathcal{M}|)$ from LLMs in $\mathcal{M}$ to obtain the final classification result $T$. Noting that $S=\{s_i\}_{i=1}^{|S|}$ and $T=\{t_i\}_{i=1}^{|S|}$ (where $s_i$ and $t_i$ are the prediction span and prediction type for the same entity), we can obtain the prediction named entity set $Y$ as Eq. \ref{eq:zip}.
\begin{equation}
\label{eq:classfication}
T_{i}=M_{i}(X,S,P(I_{\text{cls}},D_{\text{cls}}))
\end{equation}
\begin{equation}
\label{eq:vote}
T=\text{Vote}(\{T_i\}_{i=1}^{|\mathcal{M}|})
\end{equation}
\begin{equation}
\label{eq:zip}
Y=\{(s_i,t_i)|i=1,2,\ldots,|S|\}
\end{equation}

\subsection{Type verification}
\label{sec:verification}
As shown in Fig. \ref{fig:pipeline}, the span classification stage is aimed at performing self-validation for the entity types to filter out the possible noise entities introduced in the span extraction stage and erroneous type judgments in the span classification.

The span extraction stage integrates the extraction results of multiple LLMs by taking the union. Although this process covers more spans, it may also introduce some noisy spans. Moreover, even if multiple LLMs are jointly involved in deciding the type of span, misclassification is hard to avoid. Therefore, we would like to employ a LLM verifier to verify the extracted spans and their prediction types by zero-shot learning, thus filtering out noisy and misclassified cases. We pick a LLM in $\mathcal{M}$ as the LLM verifier $M^{v}$ to determine one by one whether the prediction named entities in $Y$ are correct (i.e., whether their prediction spans and their prediction types match). If $M^{v}$ determines that the prediction is correct, the prediction is retained, otherwise it will be filtered from $Y$, thus obtaining the final prediction named entities set $Y_{\text{final}}$. This process is shown as Eq. \ref{eq:verify}, where $I_{\text{ver}}$ is the instruction using to verify a single prediction entity.
\begin{equation}
\label{eq:verify}
Y_{\text{final}} = \{ y \in Y \mid M^v(y, P(I_{\text{ver}})) = \text{True} \}
\end{equation}

\section{Experiments}
In this section, we conduct a series of experiments on multiple widely adopted NER datasets from diverse domains and compare EL4NER to the most advanced ICL-based NER methods to answer the following research questions: 

\begin{itemize}[leftmargin=*]
	\item \textbf{RQ 1}: Does EL4NER outperform other compared methods based on large-parameter LLMs across various datasets?
    \item \textbf{RQ 2}: How much specific performance gain does each component in EL4NER have?
    \item \textbf{RQ 3}: How different LLM verifiers affect EL4NER performance?
    \item \textbf{RQ 4}: How does the number of backbones affect the performance of EL4NER?
    \item \textbf{RQ 5}: How does the number of demonstrations in ICL affect the performance of EL4NER?
\end{itemize}

\begin{table*}[t]
    \centering
    \resizebox{\textwidth}{!}{%
    \begin{tabular}{ccccccc}
        \hline
        \multirow{2}{*}{\textbf{Category}} & \multirow{2}{*}{\textbf{Method}} & \multirow{2}{*}{\textbf{Backbone}} & \multirow{2}{*}{\textbf{\#Parameters}} & \multicolumn{3}{c}{\textbf{Dataset}} \\
        \cline{5-7}
        & & & & ACE05 & GENIA & WNUT17 \\
        \hline
        \multirow{5}{*}{\textbf{Baselines}} & CodeIE & \multirow{5}{*}{GPT-4o} & \multirow{5}{*}{/} & 62.17 & 64.73 & 45.11 \\
        & P-ICL &  &  & 58.34 & 54.16 & 28.05 \\
        & Self-Improving &  &  & 47.94 & 58.06 & 37.43 \\
        & GPT-NER &  &  & \textbf{76.34} & \underline{68.78} & 46.97 \\
        & LT-NER &  &  & 66.16 & 68.36 & \underline{53.88} \\
        \cdashline{1-7}
        \multirow{3}{*}{\textbf{Ours}} & \multirow{3}{*}{EL4NER} & Phi-4, & \multirow{3}{*}{37B} & \multirow{3}{*}{\underline{69.32}} & \multirow{3}{*}{\textbf{69.28}} & \multirow{3}{*}{\textbf{55.33}} \\
        & & GLM-4-9B-Chat, & & & & \\
        & & Qwen-2.5-14B-Instruct & & & & \\
        \cdashline{1-7}
        \multirow{9}{*}{\textbf{Ablation}} & \multirow{3}{*}{w/ Cosine Demo Retrieval} &  & \multirow{9}{*}{37B} & \multirow{3}{*}{66.41} & \multirow{3}{*}{68.89} & \multirow{3}{*}{51.65} \\
        & &   & & & & \\
        & &  & & & & \\
        
        & \multirow{3}{*}{w/o Task Decomposition} & Phi-4, &  & \multirow{3}{*}{68.61} & \multirow{3}{*}{67.70} & \multirow{3}{*}{57.27} \\
        & &  GLM-4-9B-Chat, & & & & \\
        & &  Qwen-2.5-14B-Instruct & & & & \\
        
         & \multirow{3}{*}{w/o Type Verification} &  &  & \multirow{3}{*}{65.42} & \multirow{3}{*}{66.17} & \multirow{3}{*}{52.86} \\
        & &   & & & & \\
        & &   & & & & \\
    
        \hline
    \end{tabular}
    } 
    \caption{Performance comparisons (\%) on ACE05, GENIA, and WNUT17. The best performance between ours and the baselines is in \textbf{boldface} and the second runners between ours and the baselines are \underline{underlined}.}
    \label{tab:main}
\end{table*}

\subsection{Experimental setup} 

\paragraph{Datasets} Our test datasets come from a variety of domains, including ACE05 \citep{doddington2004automaticACE} from the journalism domain, GENIA \citep{kim2003genia} from the biomedical domain, WNUT17 \citep{wnut} from the social media domain. When testing our method on each test set, we select its corresponding training set to form the candidate set used for demonstration retrieval stage in our method. 

\paragraph{Backbones} We choose three widely used open-source small-parameter LLMs as the backbones of our method, including GLM-4-9B-Chat \citep{glm2024chatglm}, Phi-4 \citep{abdin2024phi4technicalreport}, and Qwen-2.5-14B-Instruct \citep{qwen2.5}. For the type verification stage in EL4NER's pipeline, we performed a careful analysis for the impact of different LLM verifiers on the EL4NER's performance (as shown in Section \ref{sec:verifier_analysis}), and thus set the LLM verifier to be the GLM-4-9B-Chat.

\paragraph{Implementation details for EL4NER} For the demonstration retrieval stage in EL4NER's pipeline, we uniformly set the number of demonstrations retrieved k to 20. For all ICL processes in the pipeline, we set the temperature to 0. For all ensemble learning processes in the pipeline, we use parallel techniques to allow different backbones to inference simultaneously on different GPUs to improve the overall efficiency of the method. All experiments were conducted using vLLM framework \citep{kwon2023efficient} with Python 3.12, on an Ubuntu server equipped with 8 NVIDIA GeForce RTX 3090 GPUs and an Intel(R) Xeon(R) CPU.

\paragraph{Baselines} We compared EL4NER with the most advanced ICL-based NER methods based on close-source, large-parameter LLMs: CodeIE \citep{li2023codeielargecodegeneration}, P-ICL \citep{picl}, Self-Improving \citep{SelfImprovingFZ}, GPT-NER \citep{gptner}, LT-NER \citep{yan2024ltnerlargelanguagemodel}. The detailed description and implementation settings are in Appendix~\ref{sec:Baseline Details}.

\paragraph{Metric} Following the mainstream NER experimental setup, we adopted the micro-f1 score as our evaluation metric.

\subsection{Comparison experiment (RQ 1)}
To answer RQ 1, we list the performance metrics of EL4NER and the baselines in Table~\ref{tab:main}.

Among the baselines, GPT-NER, in which demonstration retrieval and self-validation mechanisms are also introduced, demonstrate more significant advantages, especially in ACE05. Both of GPT-NER and EL4NER employ a span-level based demonstration retrieval and self-validation mechanism. However, the retrieval mechanism of GPT-NER treats each span in the input text sequence equally and only considers candidate text sequences that share the same (not similar) spans with the input text sequence. While EL4NER weights different spans in the input text sequence based on part-of-speech tags and takes into account candidate text sequences containing similar spans by matching ones in the input text sequence with spans in the candidate text sequence. In addition, EL4NER employs multi-stage ensemble learning based on task decomposition, which accomplishes the NER task at a finer granularity, thus enabling it to outperform GPT-NER on most datasets. In terms of parameter counts, all the baselines adopt gpt-4o as their backbones, which is one of the most powerful commercial closed-source large-parameter LLMs at present. While OpenAI has not revealed GPT-4o’s exact parameter count, prevailing estimates place it at no fewer than 150 billion parameters. In contrast, EL4NER adopts three open-source small-parameter LLMs as backbones, and their total parameter count is only 37B, which \textbf{greatly reduces deployment and inference costs}. In terms of effectiveness, the micro-f1 score of EL4NER on WNUT17 and GENIA \textbf{dramatically outperforms that of all compared methods}, while its micro-f1 score on ACE2005 is second only to that of GPT-NER. \textbf{Overall, EL4NER achieves excellent NER effect with a smaller parameter count, reflecting its parametric efficiency.}

\subsection{Ablation study (RQ 2)}
To answer RQ 2, we list the performance metrics of the ablation studies for the key components in EL4NER. Overall, EL4NER has three key components:

\paragraph{Customized demonstration retrieval} To ensure a fair comparison, we replaced EL4NER’s customized demonstration retrieval with a cosine similarity-based variant (Ours w/ Cosine Demo Retrieval). This resulted in lower micro-F1 scores across all datasets, especially ACE05 and WNUT17, highlighting the effectiveness of our proposed method. While semantic similarity retrieval favors contextually similar demonstrations, our method prioritizes span-level similarity—more beneficial for NER—and further weighs spans by the part-of-speech of their head word to better identify likely named entities.

\paragraph{Task decomposition} In order to explore the effectiveness of task decomposition in EL4NER, we replace the span extract stage and the span classification stage with the process of extracting named entities from the input text sequence at once, thus obtaining the variant Ours (w/o Task Decomposition). The micro-f1 score for this variant decreased on most datasets, but increased on the WNUT17 dataset. This suggests that task decomposition leads to performance gains on most cases, but the decomposition of NER into two sequential atomic tasks suffers from the potential error accumulation, which may result in performance degradation in rare cases.

\paragraph{Type verification} We obtain the Ours (w/o Type Verification) variant by removing the type verification stage from the span classification stage. The micro-f1 score of this variant has a more significant decrease on all three datasets, indicating the importance of this component. In our pipeline, the ensemble for multiple LLMs and multiple stages makes the probability of introducing noise much higher, and type verification can effectively filter the noise, resulting in a larger performance gain.

\subsection{Analysis for different LLM verifiers (RQ 3)}
\label{sec:verifier_analysis}
\begin{figure}[t]
  \centering
  \begin{subfigure}[t]{.48\linewidth}        
    \centering
    \includegraphics[width=\linewidth]{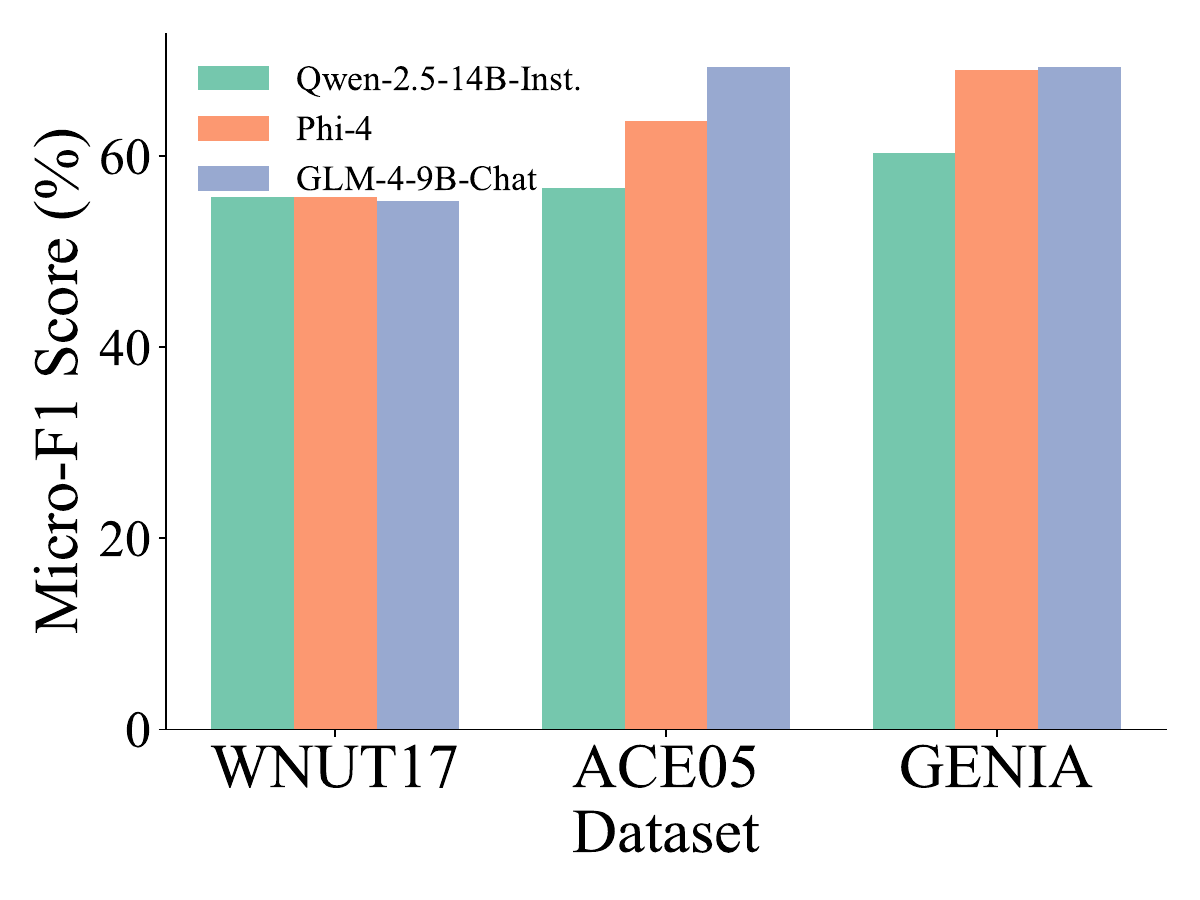}
    \caption{Micro-f1 scores of three EL4NER variants  (using Phi-4, Qwen-2.5-14B-Instruct, and GLM-4-9B-Chat as verifer, respectively) on the three datasets.}
    \label{fig:verifier_fig}
  \end{subfigure}\hfill
  \begin{subfigure}[t]{.48\linewidth}
    \centering
    \includegraphics[width=\linewidth]{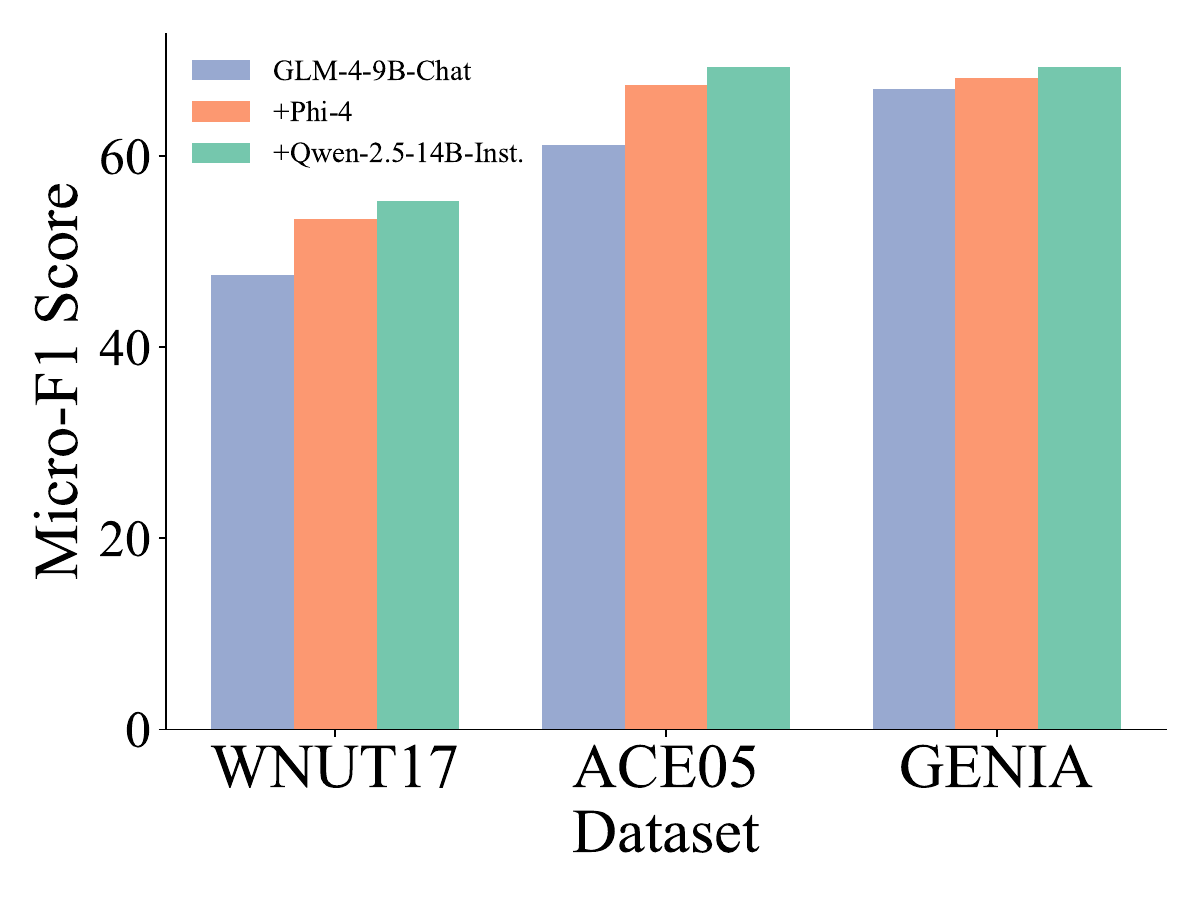}
    \caption{Micro-f1 scores of three EL4NER variants  (using one, two, and three backbones, respectively) on the three datasets.}
    \label{fig:backbone_fig}
  \end{subfigure}
  \caption{Overall comparison of variants.}
  \label{fig:verifier_backbone}
\end{figure}

In the type verification stage, we set one of the backbones to LLM verifier for self-verification. To answer RQ 3, we conduct experiments on the three datasets using each of the three backbones as an LLM verifier, and the results of the experiments are shown in Fig. \ref{fig:verifier_fig}. On the ACE05 and GENIA, Qwen-2.5-14B-Instruct performed lower, while the GLM-4-9B-Chat performed best. On WNUT17, the three backbones performed almost identically. Considering that GLM-4-9B-Chat has the lowest parameter count among the three, the experimental results suggest that certain smaller-parameter LLMs may also lead to better performance in the validation task, probably because they use more judgment-related training data in the instruction tuning phase. Based on the experimental results, we choose GLM-4-9B-Chat as the LLM verifier in EL4NER .

\subsection{Analysis for the number of backbones (RQ 4)}
To answer RQ 4, in addition to the original variant (i.e., EL4NER that employs all the three LLMs as backbones), we form two other variants of EL4NER by respectively picking one and two of the three LLMs as backbones. The performance comparison of these three variants on the three datasets is shown in Fig. \ref{fig:backbone_fig}. The performance of EL4NER on all three datasets gradually increases as the number of backbones increases, which is particularly evident on WNUT17 and ACE05, indicating that the proposed ensemble learning pipeline can bring performance gain on the NER task. In addition, in most cases, the performance gain is larger when the number of backbones is boosted from one to two, and smaller when it is boosted from two to three, which suggests that the performance gain of ensemble learning becomes progressively slower as the number of backbones increases. In ensemble learning, when the errors of individual models are compensated by those of others, the overall performance can be significantly enhanced. As the number of backbones increases, the initial additions lead to a considerable reduction in error due to the complementary error patterns among the backbones. However, when the third backbone is introduced, its predictions are likely to be highly correlated with those of the first two (i.e., the existing backbones have already captured most of the complementary information), thereby providing less additional unique information. This results in diminishing marginal returns in performance improvement.

\subsection{Analysis for the number of retrieved demonstrations in ICL (RQ 5)}
\begin{wrapfigure}{r}{0.4\textwidth}
  \centering
  \includegraphics[width=0.4\textwidth]{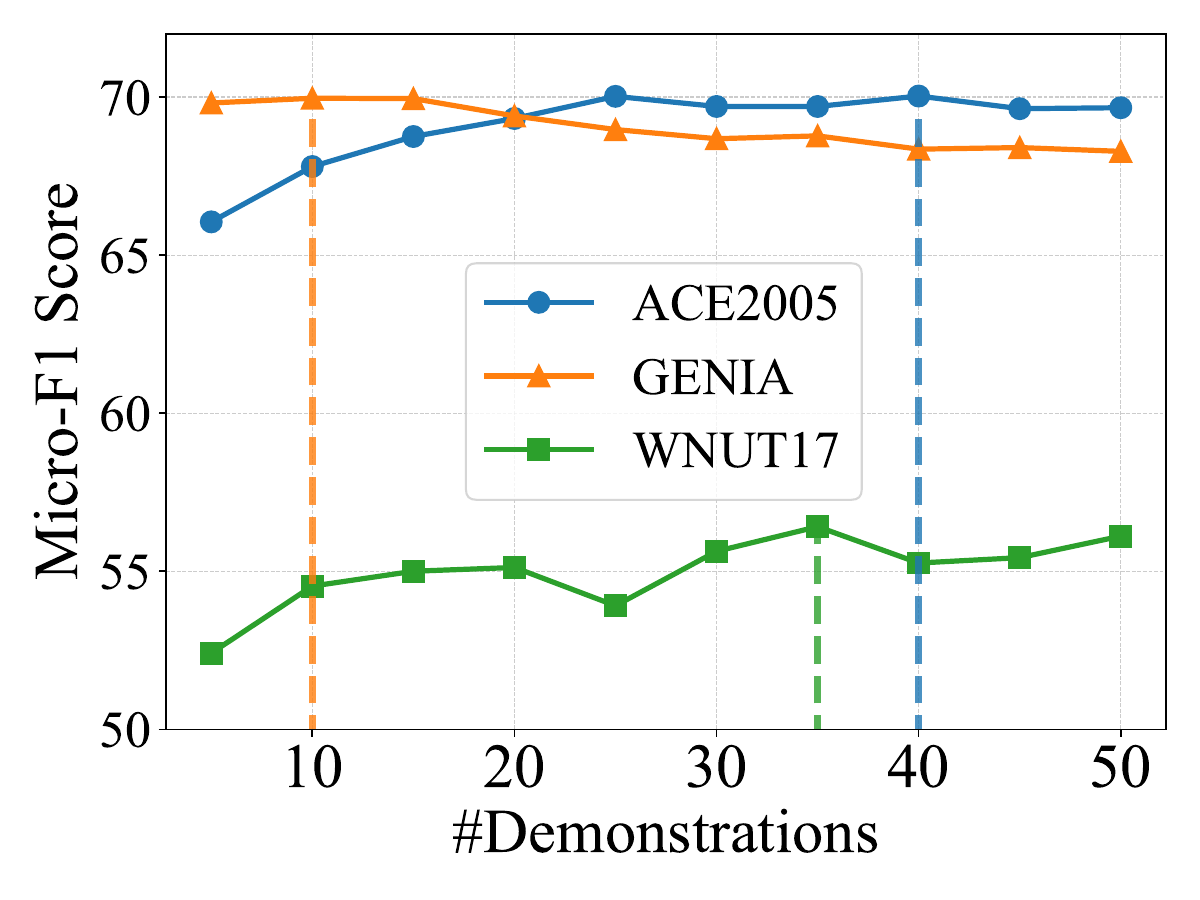}
  \caption{EL4NER performance vs.\ \#demonstrations in the three datasets.}
  \label{fig:demos}
\end{wrapfigure}
To answer RQ 5, Fig. \ref{fig:demos} plots EL4NER’s performance on three datasets as we vary the in-context learning demonstration count $k$. Three key insights arise. First, each dataset exhibits a distinct performance trajectory, indicating that not only the number but also the sequence of demonstrations in context influences results. Second, adding more demonstrations is not always get better performance, which can challenge backbones' limited context windows, cause the “lost in the middle” problem and actually harm the method's performance. Third, the optimal $k$ differs across datasets, mirroring the backbone models’ domain proficiency. For example, their strong biomedical expertise allows them to peak on GENIA with just 10 demonstrations, whereas in the news domain roughly 40 demonstrations are needed to attain best performance.

\section{Conclusion}
In this paper, we propose a lightweight and efficient NER method, called EL4NER, which performs a multi-stage and deep-level integration for NER results from multiple open-source small-parameters based on ensemble learning and task decomposition. In addition, we design a span-level demonstration retrieval mechanism based on span pre-extraction and weights of part-of-speech tags to enhance the NER effect of LLMs, and introduces a self-verification mechanism to filter the noise introduced in the ensemble process. With high parameter efficiency, EL4NER meets or even exceeds the performance of the start-of-the-art NER method using the LLMs in GPT series as the backbone on three widely used NER datasets, demonstrating its validity and underscore the feasibility of employing open-source,
small-parameter LLMs within the ICL paradigm for NER tasks.

\newpage
\bibliographystyle{plainnat}
\bibliography{references}

\newpage
\appendix
\section{Limitations}
\label{sec:limitations}
Although EL4NER adopts a novel ensemble learning pipeline based on multiple open-source, small-parameter LLMs, outperforming other methods based on closed-source, large-parameter LLMs across multiple datasets, there are still some limitations. On the one hand, the selection for backbones and LLM verifier in EL4NER relies on experience, and for different datasets, a fixed set of backbones and LLM verifier do not always achieve best effect. Therefore, a flexible selection strategy for backbones and LLM verifier needs to be explored. On the other hand, when the classification results are integrated by voting in the span classification stage, it remains to be investigated whether the weights of the votes from different LLMs need to be kept different according to their features.

\section{Prompts in EL4NER}

\noindent
\colorbox{yellow}{\textbf{[Input Text Sequence]}} Both the AMA and the Bush administration released reports this week saying out of control trial lawyers are driving doctors out of their practices all across the country .

\vspace{1em}
\noindent
\colorbox{yellow}{\textbf{[Entity Types in Schema]}} organization, person, geographical social political, vehicle, location, weapon, facility

\vspace{1em}
\noindent
\colorbox{yellow}{\textbf{[Prompt in Span Pre-Extraction]}}\\
\begin{lstlisting}
You are a Named Entity Recognition (NER) expert. Your task is to enumerate all potential entities from the given sentence with accuracy and clarity. Follow these guidelines carefully and refer carefully to the input and output examples given.

1. **Avoid Duplicates and Single-Character Entries**:
   - Do not include single-character entries unless they are **meaningful pronouns** like "I".
   - Ensure that no duplicate entities are listed.

2. **Output Format**:
   - Begin the output with `[CLS]` and end with `[SEP]`.
   - If the sentence contains no valid entities, output `[CLS][SEP]`.
   - If the sentence contains one entity, output `[CLS]entity1[SEP]`.
   - If the sentence contains multiple entities, output `[CLS]entity1[SEP]entity2[SEP]...[SEP]entityn[SEP]`.
   - Separate each entity with `[SEP]`, and ensure proper formatting.
   - Do not include any explanations, reasoning, or comments. Present all entities as plain text, without quotation marks, special character escapes (like `\`), or any additional formatting.

By following these steps, ensure that each entity listed is a true, accurately represented element from the sentence, complete with essential modifiers, and formatted precisely as specified. Additionally, leverage prior identified entities from historical dialogue to enhance accuracy and maintain contextual relevance.

**Input:**  
Both the AMA and the Bush administration released reports this week saying out of control trial lawyers are driving doctors out of their practices all across the country.
\end{lstlisting}

\vspace{1em}
\noindent

\colorbox{yellow}{\textbf{[EL4NER's Response for Span Pre-Extraction]}}
\begin{lstlisting}
[CLS]AMA[SEP]Bush administration[SEP]reports[SEP]week[SEP]lawyers[SEP]dockers[SEP]practices[SEP]country[SEP]
\end{lstlisting}

\vspace{1em}
\noindent
\colorbox{yellow}{\textbf{[Prompt in Span Extraction]}}\\

\begin{lstlisting}
You are a Named Entity Recognition (NER) expert. Your task is to enumerate all potential entities from the given sentence with accuracy and clarity. Follow these guidelines carefully and refer carefully to the input and output examples given.

1. **Avoid Duplicates and Single-Character Entries**:
   - Do not include single-character entries unless they are **meaningful pronouns** like "I".
   - Ensure that no duplicate entities are listed.
   
2. **Output Format**:
   - Begin the output with [CLS] and end with [SEP].
   - If the sentence contains no valid entities, output [CLS][SEP].
   - If the sentence contains one entity, output [CLS]entity1[SEP].
   - If the sentence contains multiple entities, output [CLS]entity1[SEP]entity2[SEP]...[SEP]entityn[SEP].
   - Separate each entity with [SEP], and ensure proper formatting.
   - Do not include any explanations, reasoning, or comments. Present all entities as plain text, without quotation marks, special character escapes (like "\"), or any additional formatting.
   
By following these steps, ensure that each entity listed is a true, accurately represented element from the sentence, complete with essential modifiers, and formatted precisely as specified. Additionally, leverage prior identified entities from historical dialogue to enhance accuracy and maintain contextual relevance.

**Input:**
During the high-profile summit in New York, the Bush administration, led by Bush, introduced major reforms that their supporters applauded, but a group of outspoken trial lawyers strongly opposed, prompting debate in Germany and France.

**Output:**
[CLS]New York[SEP]Bush administration[SEP]Bush[SEP]their[SEP]outspoken trial lawyers[SEP]Germany[SEP]France[SEP]

*...k-1 more demonstrations...*

**Input:**
Both the AMA and the Bush administration released reports this week saying out of control trial lawyers are driving doctors out of their practices all across the country .

**Output:**
\end{lstlisting}

\vspace{1em}
\noindent
\colorbox{yellow}{\textbf{[EL4NER's Response for Span Pre-Extraction]}}
\begin{lstlisting}
[CLS]AMA[SEP]Bush administration[SEP]reports[SEP]week[SEP]dockers[SEP]practices[SEP]country[SEP]Bush[SEP]their[SEP]out of control trial lawyers[SEP]
\end{lstlisting}

\vspace{1em}
\noindent
\colorbox{yellow}{\textbf{[Prompt in Span Classification]}}\\
\begin{lstlisting}
You are a Named Entity Recognition (NER) expert. Your task is to classify only the given potential entities in the provided sentence. Follow these instructions precisely:
1. **Input Format**:
   - The input will be presented in the format: "[CLS]sentence[SEP]entity1[SEP]entity2".
   - Classify **only the provided entities** (e.g., entity1, entity2) exactly as given. **Do not introduce any new entities** that are not explicitly listed.
2. **Output Format**:
   - For each entity, output the classification in the following format: "[CLS]entity1[SEP]type1[CLS]entity2[SEP]type2".
   - Ensure each classification strictly follows this format, with no added explanations, reasoning, or comments.
3. **Classification Rules**:
   - Classify each entity into one of the specified types: organization, person, geographical social political, vehicle, location, weapon, facility.
   - Adhere to the **exact casing** of each type as provided, and remember that each entity can belong to only one type.
4. **Output Requirements**:
   - Ensure all entities and types are output as plain text without quotation marks, special character escapes (like "\"), or any additional formatting.
   - Confirm each entity listed in the input is classified; no omissions or additions should occur.
By following these steps, accurately classify each provided entity based on the specified types and ensure a consistent format in your output.

**Input:** [CLS]During the high-profile summit in New York, the Bush administration, led by Bush, introduced major reforms that their supporters applauded, but a group of outspoken trial lawyers strongly opposed, prompting debate in Germany and France.[SEP]New York[SEP]Bush administration[SEP]Bush[SEP]their[SEP]outspoken trial lawyers[SEP]Germany[SEP]France
**Output:** [CLS]New York[SEP]geographical social political[CLS]Bush administration[SEP]organization[CLS]Bush[SEP]person[CLS]their[SEP]person[CLS]outspoken trial lawyers[SEP]person[CLS]Germany[SEP]geographical social political[CLS]France[SEP]geographical social political
*...k-1 more demonstrations...*
**Input:** [CLS]Both the AMA and the Bush administration released reports this week saying out of control trial lawyers are driving doctors out of their practices all across the country .[SEP]AMA[SEP]Bush administration[SEP]reports[SEP]week[SEP]dockers[SEP]practices[SEP]country[SEP]Bush[SEP]their[SEP]out of control trial lawyers

**Output:**
\end{lstlisting}

\vspace{1em}
\noindent
\colorbox{yellow}{\textbf{[EL4NER's Response for Span Extraction]}}
\begin{lstlisting}
[CLS]AMA[SEP]organization[CLS]Bush administration[SEP]organization[CLS]reports[SEP]organization[CLS]week[SEP]organization[CLS]dockers[SEP]person[CLS]practices[SEP]facility[CLS]country[SEP]geographical social political[CLS]Bush[SEP]person[CLS]their[SEP]person[CLS]out of control trial lawyers[SEP]person
\end{lstlisting}

\vspace{1em}
\noindent
\colorbox{yellow}{\textbf{[Prompt in Type Verification]}}\\
\begin{lstlisting}
You are a highly skilled linguist specializing in Named Entity Recognition (NER) tasks. Your task is to carefully evaluate whether the entity "reports" in the sentence "Both the AMA and the Bush administration released reports this week saying out of control trial lawyers are driving doctors out of their practices all across the country ." corresponds to the type "organization". Follow these guidelines:

1. **Review Modifiers and Prepositions**:
   - Pay special attention to any **modifiers or descriptors** directly preceding the noun. Be cautious with **prepositional phrases**, as they often do not constitute valid entities on their own (e.g., "all over the world" should generally not be marked as an entity).
   - Pay close attention to any **modifiers, descriptors, or titles** directly preceding the noun. Ensure that these elements are included as part of the entity when relevant, as entities without their associated modifiers may be incomplete and invalid. For instance, if the sentence contains "President Biden," "Biden" without "President" should not be considered a valid entity in this context.
2. **Examine Pronouns**:
   - Only consider **personal pronouns** and **possessive pronouns** (e.g., "they", "their", "us") as potential entities if they clearly refer to people or defined entities.
   - **Ignore** other types of pronouns, such as demonstrative pronouns (e.g., "this", "that") and indefinite pronouns (e.g., "both", "some"), as these are typically not valid entities.
3. **Output Only True or False**:
   - If "reports" corresponds to the type "organization", output **true**.
   - If it does not correspond, output **false**.
Respond strictly with "true" or "false", without any additional explanation or reasoning.
\end{lstlisting}

\vspace{1em}
\noindent
\colorbox{yellow}{\textbf{[EL4NER's Response for Type Verification]}}
\begin{lstlisting}
false
\end{lstlisting}

\section{Experiment Details}
\subsection{Datasets Details}
We employ three widely used NER datasets from multiple domains in our experiments, which are ACE2005, GENIA, and WNUT17. The details of their background information are described below.

(1) The \textbf{ACE05} dataset is a corpus developed under the Advanced Content Extraction (ACE) program by the U.S. Department of Defense. It is widely utilized for tasks such as named entity recognition, relation extraction, and event detection. In the context of NER, the dataset includes a diverse range of entity categories, such as person names, organization names, and geographical locations, with samples drawn from multiple sources including news reports and broadcast transcripts. Owing to its rigorous annotation standards and heterogeneous text sources, ACE05 serves as a critical benchmark for evaluating entity recognition algorithms.

(2) The \textbf{GENIA} dataset is a cornerstone resource in the biomedical domain, primarily derived from biomedical literature and abstracts. It offers detailed annotations for domain-specific entities, including proteins, DNA, RNA, and cell types, reflecting the unique knowledge structure and specialized terminology inherent to biomedicine. This high-quality annotated corpus provides robust support for biomedical named entity recognition tasks and related natural language processing research, thereby facilitating advancements in the field.

(3) Originating from the 2017 WNUT shared task, the \textbf{WNUT17} dataset focuses on the challenges of identifying novel and rare entities within informal texts such as social media content. The dataset comprises a large collection of user-generated texts, including tweets and forum posts, which are characterized by their informal language style and inherent noise. Consequently, WNUT17 offers a valuable testbed for evaluating the performance of NER systems under the dynamic and unstructured conditions typical of real-world social media environments.

Details statistics of datasets can be found in Table \ref{tb:statistics}.

\begin{table}[ht]
\centering
\resizebox{\textwidth}{!}{%
    \begin{tabular}{c c c c c c}
    \hline
    \textbf{Dataset} & \textbf{Domain} & \textbf{\#Entity Types} & \textbf{\#Train} & \textbf{\#Dev} & \textbf{\#Test} \\
    \hline
    ACE05 (Walker et al., 2006) & News          & 7 & 7299  & 971  & 1060 \\
    GENIA (Kim et al., 2003)    & Biomedical    & 5 & 15023 & 1669 & 1854 \\
    WNUT17 (Derczynski et al., 2017) & Social Media & 6 & 3394  & 1009 & 1287 \\
    \hline
    \end{tabular}
}
\caption{The statistics of datasets.}
\label{tb:statistics}
\end{table}

\subsection{Backbones Details}
In EL4NER, we mainly employ three open-source small-parameter LLMs as backbones, which are GLM-4-9B-Chat, Phi-4 and Qwen2.5-14B-Instruct. The details of these LLMs are as follows.

(1) \textbf{GLM-4-9B-Chat} is an open-source model from the GLM-4 series developed by Zhipu AI, featuring 9B parameters. It demonstrates high performance across various tasks, including semantic understanding, mathematical reasoning, code generation, and knowledge-based question answering. Beyond supporting multi-turn dialogues, GLM-4-9B-Chat offers advanced functionalities such as web browsing, code execution, custom function calls, and long-text reasoning with a context length of up to 128K tokens. Additionally, this model supports 26 languages, including Japanese, Korean, and German.

(2) \textbf{Phi-4} is a small-scale language model introduced by Microsoft, comprising 14B parameters. It employs a decoder-only Transformer architecture, initially supporting a context length of 4,096 tokens, which was later extended to 16,000 tokens during mid-training phases. The training regimen emphasizes high-quality synthetic data, generated through techniques like multi-agent prompting and self-refinement workflows, focusing on diversity, complexity, and accuracy. Phi-4 exhibits performance in mathematical reasoning and problem-solving tasks comparable to larger models.

(3) \textbf{Qwen2.5-14B-Instruct} is a 14B parameter large language model developed by Alibaba Cloud's Tongyi Qianwen team. Built upon the Transformer architecture, it incorporates techniques such as Rotary Position Embeddings (RoPE), SwiGLU activation, and RMSNorm normalization, facilitating causal language modeling. The model supports a context length of up to 131,072 tokens and excels across various tasks. Furthermore, it demonstrates strong performance in multilingual understanding and generation, making it suitable for diverse natural language processing applications.

\subsection{Baseline Details} \label{sec:Baseline Details}
To ensure the advanced capabilities of LLMs and align with other baseline settings, we use \texttt{GPT-4o-2024-11-20} as the base model for all baselines.

\textbf{CodeIE} \citep{li2023codeielargecodegeneration}  
reformulates NER tasks into code generation problems, leveraging the structured nature of code to align with the pre-training of code generation models and enhancing performance in extracting structured information from text. In the original CodeIE, the Code-LLM used was \texttt{code-davinci-002}, which is Codex from OpenAI. Codex is a large language model adapted from \texttt{GPT-3} and further pre-trained on open-source codebases. The \texttt{code-davinci-002} version of Codex supports up to 8k input tokens. Since this API is no longer accessible through OpenAI and has become outdated, we adopted \texttt{GPT-4o-2024-11-20}, which has stronger capabilities in both code and general tasks, as the Code-LLM. Regarding the number of examples used, we maintained the original paper's 25 retrieval samples and 5 in-context examples.

\textbf{P-ICL} \citep{picl} introduces a novel perspective to NER by treating entities as anchored points within the text and using surrounding context to infer their types, effectively reducing reliance on sequence-level modeling and enabling more efficient and flexible use of large language models under few-shot scenarios. For the example setup, we used the 10+20-shot setting from the original paper, where there are 20 in-context examples, and each entity type has 10 entity examples.

\textbf{LT-NER} \citep{yan2024ltnerlargelanguagemodel} proposes a new tagging framework for NER by embedding entity candidates directly into the input with contextual markers, guiding large language models to make label predictions based on enriched local cues rather than implicit span detection, thereby enhancing precision across diverse domains. For the example setup, we used the unique 30-shot setting from the original paper, where there are 30 in-context examples.

\textbf{Self-Improving} \citep{SelfImprovingFZ} enhances zero-shot NER by leveraging large language models to automatically generate pseudo-labels and refine the model through iterative self-feedback, improving performance without annotated data. It is a truly zero-shot task that uses LLMs to perform pre-labeling in order to obtain pseudo-examples. For the number of pseudo-examples, we followed the original setting with 50 sampled retrieved examples, of which 16 are in-context examples.

\textbf{GPT-NER}  \citep{gptner}
adopts LLM of GPT series as its backbone, transforming the sequence labeling task into a text generation task with carefully designed prompts, and incorporates retrieval and self-validation mechanisms to improve its performance. In both the generation and self-verification stages, we used entity-level retrieval method, with the retrieval sample size being the full training set, and the in-context sample size being 16.

\newpage

\end{document}